\documentclass[10pt,twocolumn,letterpaper]{article}
\usepackage{cvpr}
\usepackage{times}
\usepackage{epsfig}
\usepackage{graphicx}
\usepackage{amsmath}
\usepackage{amssymb}
\usepackage{subfig}
\usepackage{authblk}
\usepackage{cite}
\usepackage{comment}
\usepackage{adjustbox}
\usepackage{textcomp}
\usepackage{array,multirow,graphicx}
\usepackage[table]{xcolor}
\usepackage{colortbl}
\usepackage{bbm}
\usepackage{makecell}
\usepackage[percent]{overpic}

\usepackage{booktabs}
\usepackage{arydshln}

\usepackage[pagebackref=true,breaklinks=true,letterpaper=true,colorlinks,bookmarks=false]{hyperref}

\cvprfinalcopy 

\def\cvprPaperID{76} 

\definecolor{TableRed}{HTML}{800000}
\newcommand{\TextRed}[1]{\textcolor{TableRed}{#1}}

\makeatletter
\newcommand\footnoteref[1]{\protected@xdef\@thefnmark{\ref{#1}}\@footnotemark}
\makeatother

\def\best{\bf \cellcolor[gray]{0.85}}
\def\secbest{\cellcolor[gray]{0.92} }

\begin{document}


\title{Context Encoding for Semantic Segmentation}

\author[1,2]{
Hang Zhang
}
\author[1]{
\; Kristin Dana
}
\author[3]{
\; Jianping Shi
}
\author[2]{
\; Zhongyue Zhang
}

\author[4]{
\\Xiaogang Wang
}
\author[2]{
\;Ambrish Tyagi
}
\author[2]{
\;Amit  Agrawal
}

\affil[ ]{$^1$Rutgers University\; $^2$Amazon Inc\; $^3$SenseTime\; $^4$The Chinese University of Hong Kong}
\affil[ ]{ 
\tt\small \{zhang.hang@,kdana@ece.\}rutgers.edu, shijianping@sensetime.com}
\affil[ ]{
\tt\small
xgwang@ee.cuhk.edu.hk, \{zhongyue,ambrisht,aaagrawa\}@amazon.com\\
}
\renewcommand\Authsep{  } 
\renewcommand\Authands{  }

\maketitle
\begin{abstract}

Recent work has made significant progress in improving spatial resolution for pixelwise labeling with Fully Convolutional Network (FCN) framework by employing Dilated/Atrous convolution, utilizing multi-scale features and refining boundaries. 
In this paper, we explore the impact of global contextual information in semantic segmentation by introducing the Context Encoding Module, which captures the semantic context of scenes and selectively highlights  class-dependent featuremaps.
The proposed Context Encoding Module significantly improves semantic segmentation results with only marginal extra computation cost over FCN. 
Our approach has achieved new state-of-the-art results 51.7\% mIoU on PASCAL-Context, 85.9\% mIoU on PASCAL VOC 2012.  
{Our single model achieves a final score of 0.5567 on ADE20K test set, which surpasses the winning entry of COCO-Place Challenge 2017}. 
In addition, we also explore how the Context Encoding Module can improve the feature representation of relatively shallow networks for the image classification on CIFAR-10 dataset. 
Our 14 layer network has achieved an error rate of $3.45\%$, which is comparable with state-of-the-art approaches with over $10\times$ more layers.
The source code for the complete system are publicly available\footnote{Links can be found at \url{http://hangzh.com/}}. 

\end{abstract}

\section{Introduction}

\begin{figure}
\centering
\includegraphics[width=\linewidth, height=0.4\linewidth]{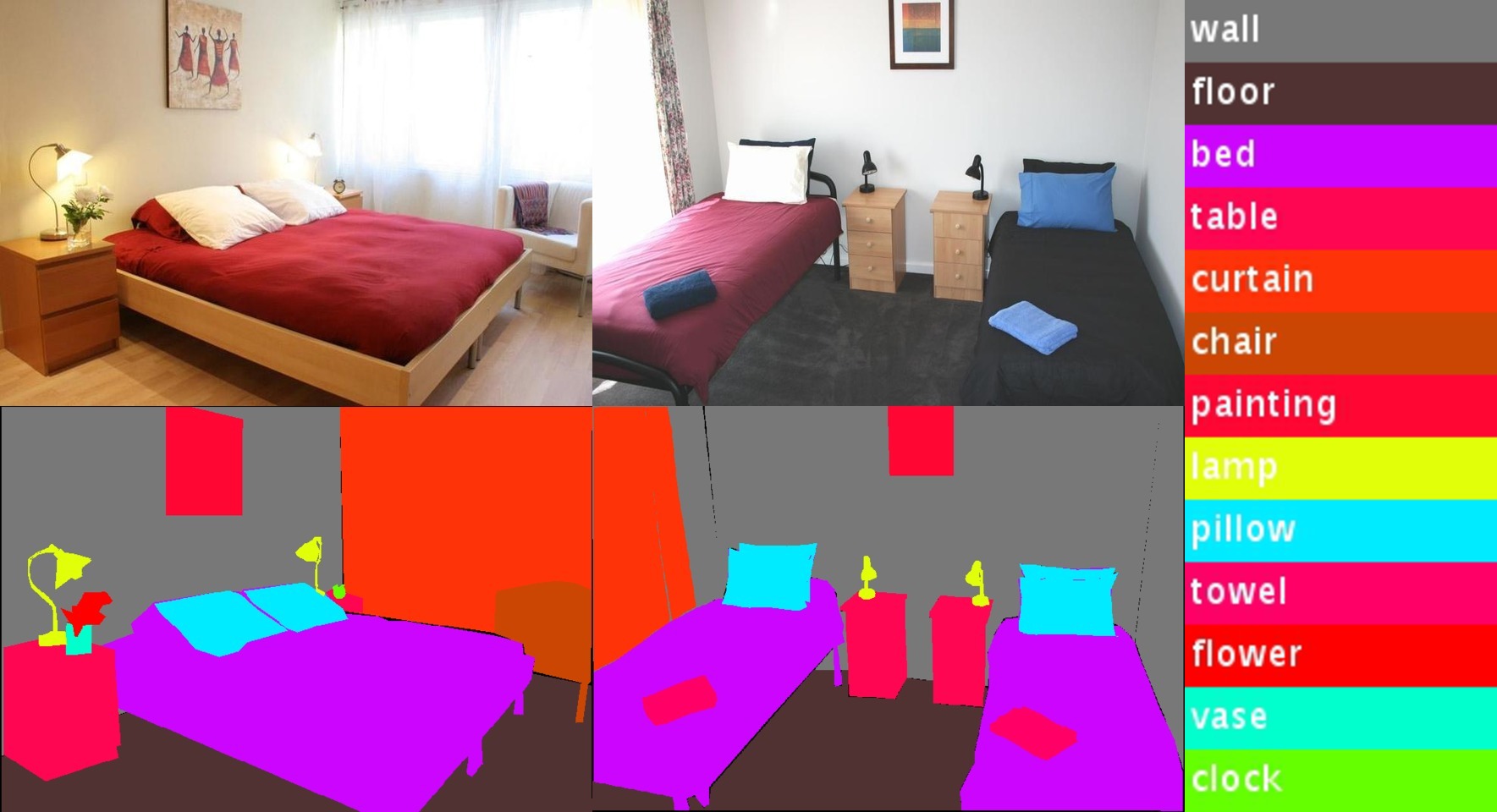}
\includegraphics[width=\linewidth, height=0.4\linewidth]{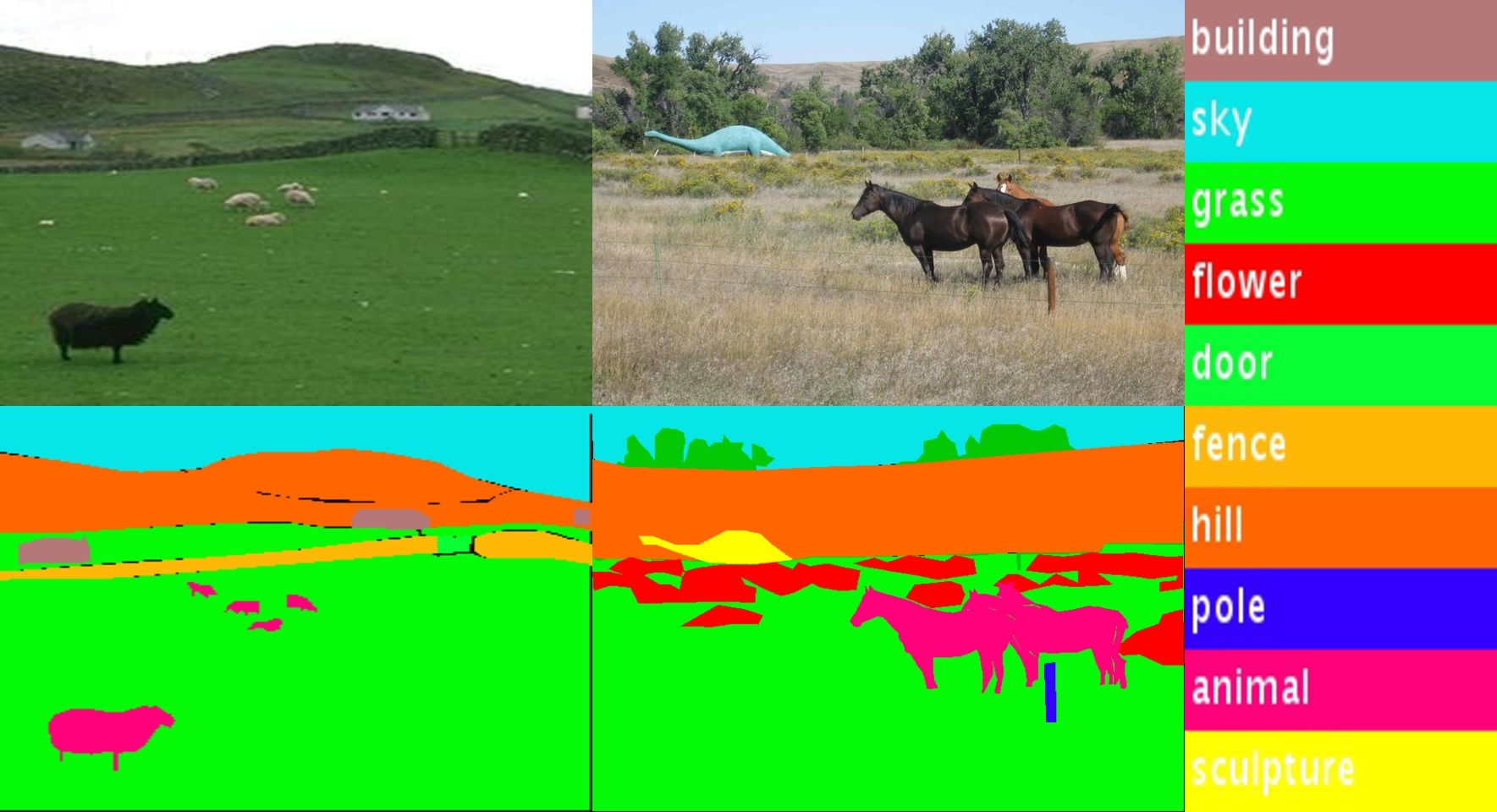}
\caption{ Labeling a scene with accurate per-pixel labels is a challenge for semantic segmentation algorithms. Even humans find the task challenging. However, narrowing the list of probable categories based on scene context makes labeling much easier.  Motivated by this, we introduce the Context Encoding Module which selectively highlights the class-dependent featuremaps and makes the semantic segmentation easier for the network. (Examples from ADE20K~\cite{zhou2017scene}.) }
\vspace{-1em}
\label{fig:abstract}
\end{figure}

Semantic segmentation assigns per-pixel predictions of  object categories for the given image, which provides a comprehensive scene description including the information of object category, location and shape. 
State-of-the-art semantic segmentation approaches are typically based on the Fully Convolutional Network (FCN) framework~\cite{long2015fully}. 
The adaption of Deep Convolutional Neural Networks (CNNs)~\cite{lecun1998gradient} benefits from the rich information of object categories and scene semantics learned from diverse set of images~\cite{imagenet}. 
CNNs are able to capture the informative representations with global receptive fields by stacking convolutional layers with non-linearities and downsampling. 
For conquering the problem of  spatial resolution loss associated with downsampling, recent work uses Dilated/Atrous convolution strategy to produce dense predictions from pre-trained networks~\cite{chen2014semantic,yu2015multi}. 
However, this strategy also isolates the pixels from the global scene context, leading to  misclassified pixels. For example in the 3$^{rd}$ row of Figure~\ref{fig:aderesults}, the baseline approach classifies some pixels in the {\it windowpane} as {\it door}.

Recent methods have achieved state-of-the-art performance by enlarging the receptive field using multi-resolution pyramid-based representations.   For example, PSPNet adopts Spatial Pyramid Pooling that pools the featuremaps into different sizes and concatenates them the after upsampling \cite{zhao2016pyramid} and Deeplab proposes an Atrous Spatial Pyramid Pooling that employs large rate dilated/atrous convolutions ~\cite{chen2016Deeplab}. While these approaches do improve performance, the context representations are not explicit, leading to the questions: {\it Is capturing contextual information the same as increasing the receptive field size? } 
Consider labeling a new image for a large dataset (such as ADE20K~\cite{zhou2017scene} containing 150 categories) as shown in Figure~\ref{fig:abstract}. 
Suppose we have a tool allowing the annotator to first select the semantic context of the image, (e.g.\ a bedroom). 
Then, the tool could provide a much smaller sublist of relevant categories (e.g.\ bed, chair, etc.), 
which would dramatically reduce the search space of possible categories. 
Similarly, if we can design an approach to fully utilize the strong correlation between scene context and the probabilities of categories, the semantic segmentation becomes easier for the network.

Classic computer vision approaches have the advantage of capturing semantic context of the scene. 
For a given input image, hand-engineered features are densely extracted using SIFT~\cite{lowe2004distinctive} or filter bank responses~\cite{leung2001representing,varma2002classifying}. 
Then a visual vocabulary (dictionary) is often learned and the global feature statistics are described by classic encoders such as Bag-of-Words (BoW)~\cite{joachims1998text,csurka2004visual,sivic2005discovering,fei2005bayesian}, VLAD~\cite{jegou2010aggregating} or Fisher Vector~\cite{perronnin2010improving}. 
The classic representations encode  global contextual information by capturing feature statistics. While the hand-crafted feature were improved greatly by CNN methods, the overall encoding process of traditional methods was convenient and powerful. 
Can we leverage the context encoding of classic approaches with the power of deep learning?
Recent work has made great progress in generalizing traditional encoders in a CNN framework~\cite{netvlad16,zhang17}. 
Zhang {\it et al.} introduces an Encoding Layer that integrates the entire dictionary learning and residual encoding pipeline into a single CNN layer to capture  orderless representations. This method has achieved state-of-the-art results on texture classification~\cite{zhang17}. 
In this work, we extend the Encoding Layer to capture global feature statistics for understanding semantic context. 

\begin{figure*}
\includegraphics[width=\linewidth]{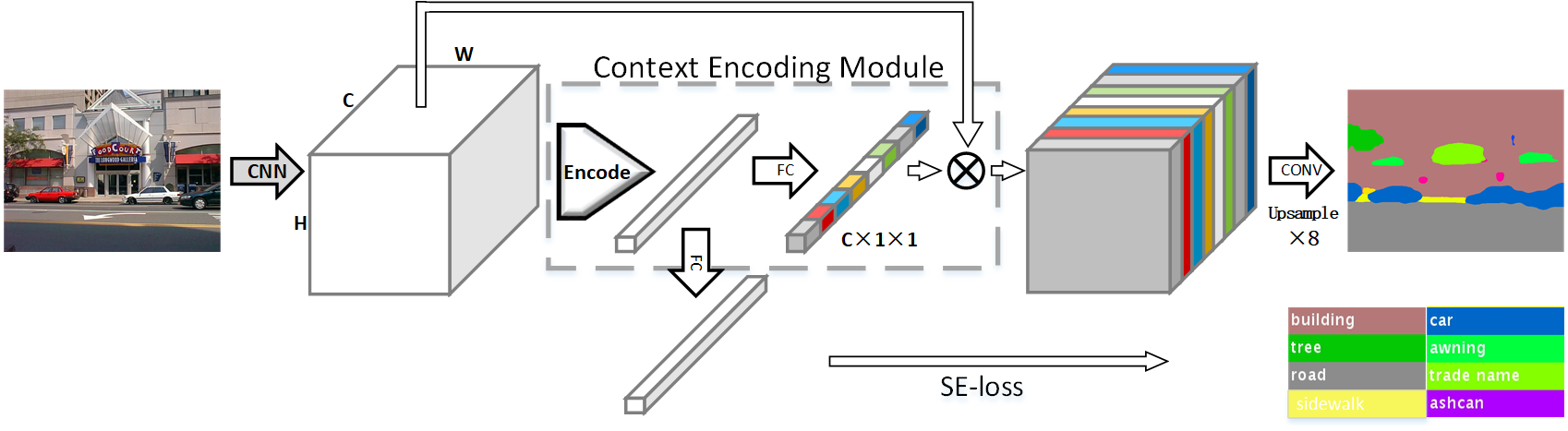}
\caption{Overview of the proposed EncNet.  Given an input image, we first use a pre-trained CNN to extract dense convolutional featuremaps. 
We build a Context Encoding Module on top, including an Encoding Layer to capture the encoded semantics and predict scaling factors that are conditional on these encoded semantics. These learned factors selectively highlight class-dependent featuremaps (visualized in colors). In another branch, we employ Semantic Encoding Loss (SE-loss) to regularize the training which lets the Context Encoding Module  predict the presence of the categories in the scene. Finally, the representation of Context Encoding Module is fed into the last convolutional layer to make per-pixel prediction. (Notation: {\it FC} fully connected layer, {\it Conv} convolutional layer, {\it Encode} Encoding Layer~\cite{zhang17}, $\bigotimes$ channel-wise multiplication.)}
\vspace{-0.8em}
\label{fig:overview}
\end{figure*}

As the {\bf first contribution} of this paper, we introduce a {\it Context Encoding Module} incorporating {\it Semantic Encoding Loss (SE-loss)}, a simple unit to leverage the global scene context information. 
The Context Encoding Module integrates an Encoding Layer to capture global context and selectively highlight the class-dependent featuremaps. 
For intuition, consider that we would want to de-emphasize the probability of a vehicle to appear in an indoor scene. 
Standard training process only employs per-pixel segmentation loss, which does not strongly utilize global context of the scene. 
We introduce Semantic Encoding Loss (SE-loss) to regularize the training, which lets the network predict the presence of the object categories in the scene to enforce  network learning of semantic context. 
Unlike per-pixel loss, SE-loss gives an equal contributions for both big and small objects and we find the performance of small objects are often improved in practice. 
The proposed Context Encoding Module and Semantic Encoding Loss are conceptually straight-forward and compatible with existing FCN based approaches.

The {\bf second contribution} of this paper is the design and implementation of a new semantic segmentation framework {\it Context Encoding Network (EncNet)}. 
EncNet augments a pre-trained Deep Residual Network (ResNet)~\cite{he2015deep} by including a Context Encoding Module as shown in Figure~\ref{fig:overview}.  
We use dilation strategy~\cite{yu2015multi,chen2014semantic} of pre-trained networks. 
The proposed Context Encoding Network achieves state-of-the-art results 85.9\% mIoU on PASCAL VOC 2012 and 51.7\% on PASCAL in Context.
Our single model of EncNet-101 has achieved a score of 0.5567 which surpass the winning entry of COCO-Place Challenge 2017~\cite{zhou2017scene}. 
In addition to semantic segmentation, we also study the power of our Context Encoding Module for visual recognition on CIFAR-10 dataset~\cite{cifar} and the performance of shallow network is significantly improved using the proposed Context Encoding Module. Our network has achieved an error rate of $3.96\%$ using only $3.5M$ parameters. 
We release the complete system including state-of-the-art approaches together with our implementation of synchronized multi-GPU Batch Normalization~\cite{ioffe2015batch} and memory-efficient Encoding Layer~\cite{zhang17}. 

\begin{figure}
\includegraphics[width=\linewidth]{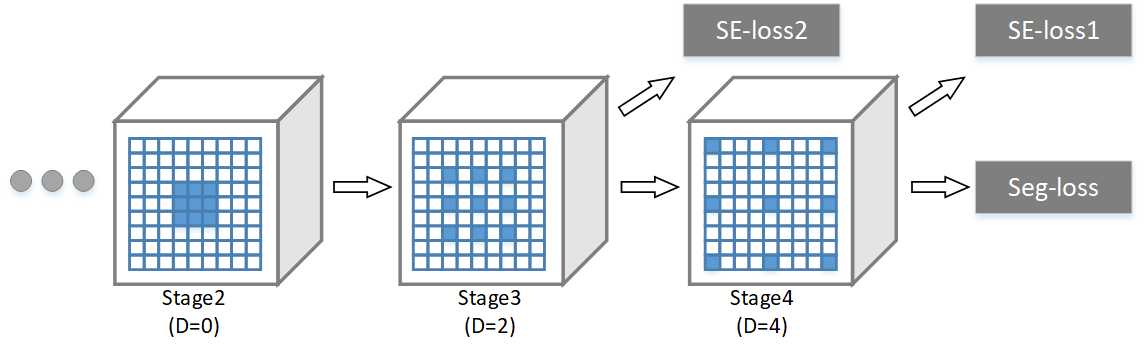}
\caption{Dilation strategy and losses. Each cube denotes different network stages. We apply dilation strategy to the stage 3 and 4. The Semantic Encoding Losses (SE-loss) are added to both stage 3 and 4  of the base network. (D denotes the dilation rate,  Seg-loss represents the per-pixel segmentation loss.) }
\label{fig:network}
\vspace{-0.8em}
\end{figure}

\section{Context Encoding Module}
We refer to the new CNN module as {\it Context Encoding Module} and the components of the module are illustrated in Figure~\ref{fig:overview}.
\paragraph{\bf Context Encoding }
Understanding and utilizing contextual information is very important for semantic segmentation. 
For a network pre-trained on a diverse set of images~\cite{imagenet}, the featuremaps encode rich information what objects are in the scene. 
We employ the Encoding Layer~\cite{zhang17} to capture the feature statistics as a global semantic context. We refer to the output of Encoding Layer as {\it encoded semantics}. 
For utilizing the context, a set of scaling factors are predicted to selectively highlight the class-dependent featuremaps.
The Encoding Layer learns an inherent dictionary carrying the semantic context of the dataset and outputs the residual encoders with rich contextual information. 
We briefly describe the prior work of Encoding Layer for completeness. 

Encoding Layer considers an input featuremap with the shape of $C\times H\times W$ as a set of $C$-dimensional input features $X=\{x_1,...x_N\}$, where $N$ is total number of features given by $H\times W$, which learns an inherent codebook $D=\{d_1,...d_K\}$ containing $K$ number of codewords (visual centers) and a set of smoothing factor of the visual centers $S=\{s_1,...s_K\}$. Encoding Layer outputs the residual encoder by aggregating the residuals with soft-assignment weights $e_k=\sum_{i=1}^Ne_{ik}$, where
\begin{equation}
e_{ik}=\frac{exp(-s_k\|r_{ik}\|^2)}{\sum_{j=1}^Kexp(-s_j\|r_{ij}\|^2)}r_{ik} ,
\end{equation}
and the residuals are given by $r_{ik}=x_i-d_k$. 
We apply aggregation to the encoders instead of concatenation. That is, $e=\sum_{k=1}^K \phi(e_k)$, where $\phi$ denotes Batch Normalization with ReLU activation, avoid making $K$ independent encoders to be ordered and also reduce the dimensionality of the feature representations. 

\paragraph{\bf Featuremap Attention } To make  use of the encoded semantics captured by Encoding Layer, we  predict scaling factors of featuremaps as a feedback loop to  emphasize or de-emphasize class-dependent featuremaps. We use a fully connected layer on top of the Encoding Layer and a sigmoid as the activation function, which outputs predicted featuremap scaling factors $\gamma = \delta(We)$, where $W$ denotes the layer weights and $\delta$ is the sigmoid function. Then the module output is given by $Y=X\otimes \gamma$ a channel wise multiplication $\otimes$ between input featuremaps $X$ and scaling factor $\gamma$.
This feedback strategy is inspired by prior work in style transfer \cite{zhang2017multistyle,huang2017adain} and a recent work SE-Net~\cite{hu2017squeeze} that tune featuremap scale or statistics. 
As an intuitive example of the utility of the approach, consider emphasizing the probability of an airplane in a sky scene, but de-emphasizing that of a vehicle. 

\paragraph{\bf Semantic Encoding Loss }
In standard training process of semantic segmentation, the network is learned from isolated pixels (per-pixel cross-entropy loss for given input image and ground truth labels). The network may have difficulty  understanding context  without global information. 
To regularize the training of Context Encoding Module, we introduce {\it Semantic Encoding Loss (SE-loss)} which forces the network to understand the global semantic information with very small extra computation cost. 
We build an additional fully connected layer with a sigmoid activation function on top of the Encoding Layer to make individual predictions for the presences of object categories in the scene and learn with binary cross entropy loss. 
Unlike per-pixel loss, SE-loss considers  big and small objects equally.
In practice, we find the segmentation of small objects are often improved.
In summary, the Context Encoding Module shown in Figure~\ref{fig:overview} captures the semantic context to predict a set of scaling factors that selectively highlights the class-dependent featuremap for semantic segmentation. 


\subsection{Context Encoding Network (EncNet)}

\begin{figure*}
\begin{minipage}[b]{0.82\linewidth}
\centering
\subfloat{
\includegraphics[width=0.25\linewidth, height=0.18\linewidth]{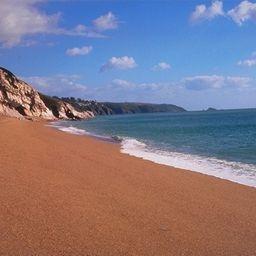}
}
\subfloat{
\includegraphics[width=0.25\linewidth, height=0.18\linewidth]{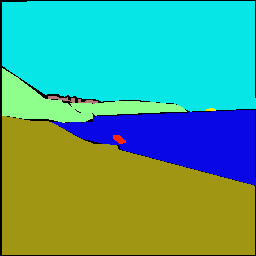}
}
\subfloat{
\includegraphics[width=0.25\linewidth, height=0.18\linewidth]{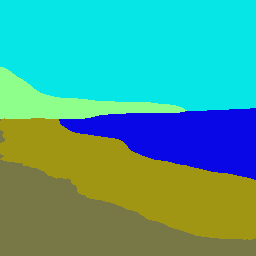}
}
\subfloat{
\includegraphics[width=0.25\linewidth, height=0.18\linewidth]{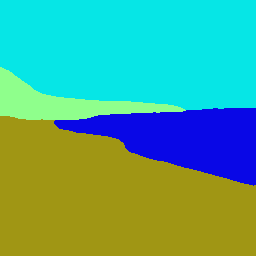}
}\\
\vspace{-0.7em}
\subfloat{
\includegraphics[width=0.25\linewidth]{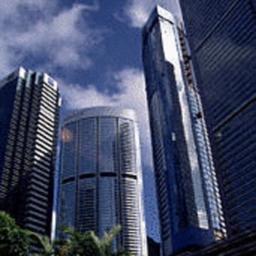}
}
\subfloat{
\includegraphics[width=0.25\linewidth]{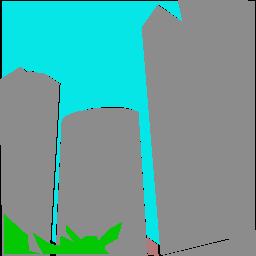}
}
\subfloat{
\includegraphics[width=0.25\linewidth]{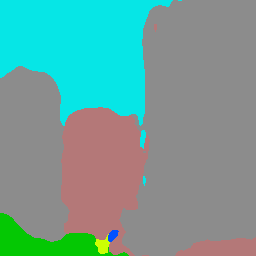}
}
\subfloat{
\includegraphics[width=0.25\linewidth]{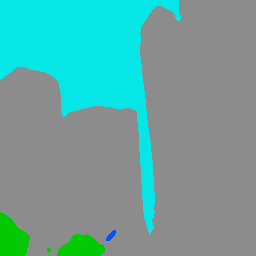}
}\\
\vspace{-0.7em}
\subfloat{
\includegraphics[width=0.25\linewidth]{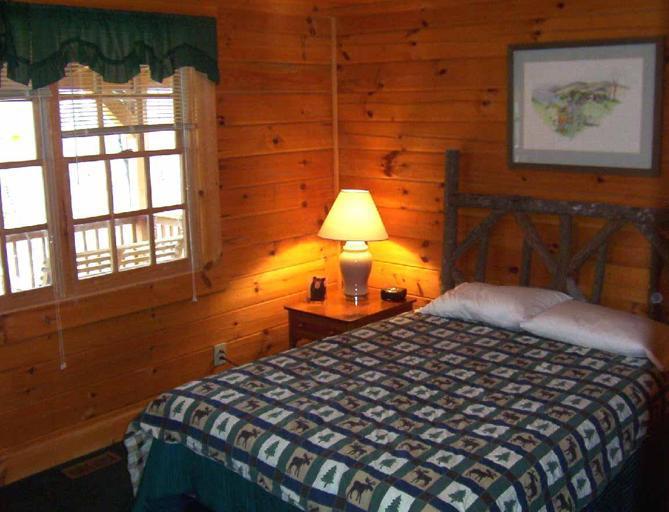}
}
\subfloat{
\includegraphics[width=0.25\linewidth]{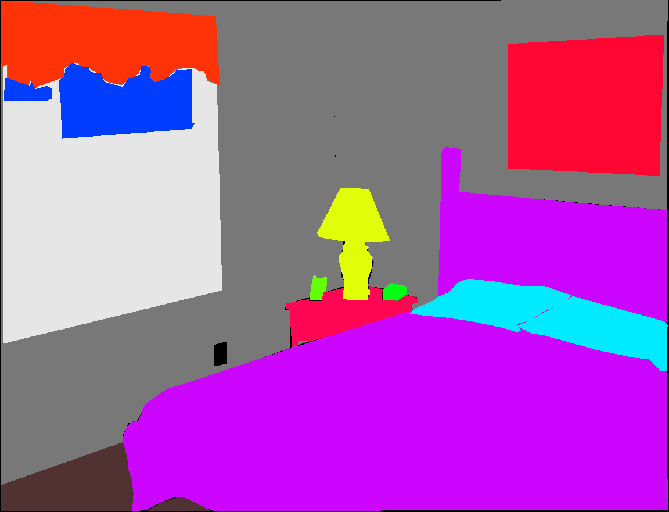}
}
\subfloat{
\includegraphics[width=0.25\linewidth]{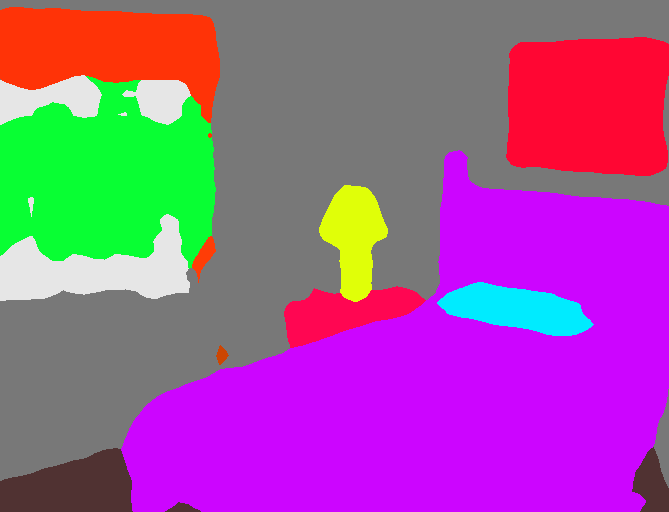}
}
\subfloat{
\includegraphics[width=0.25\linewidth]{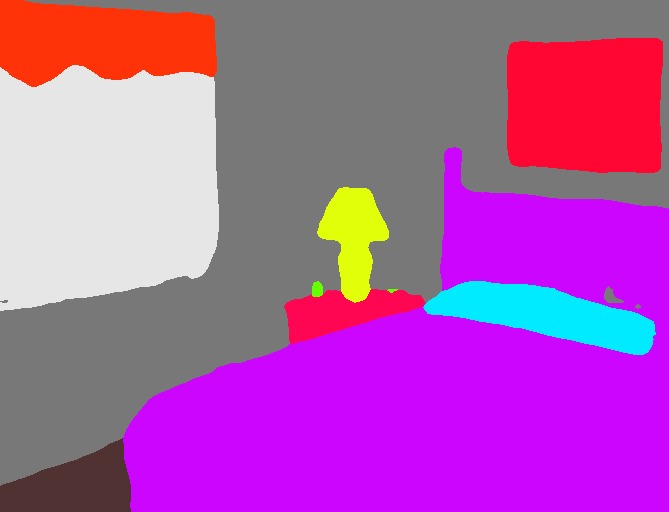}
}\\
\vspace{-0.7em}
\setcounter{subfigure}{0}
\subfloat[Image]{
\includegraphics[width=0.25\linewidth]{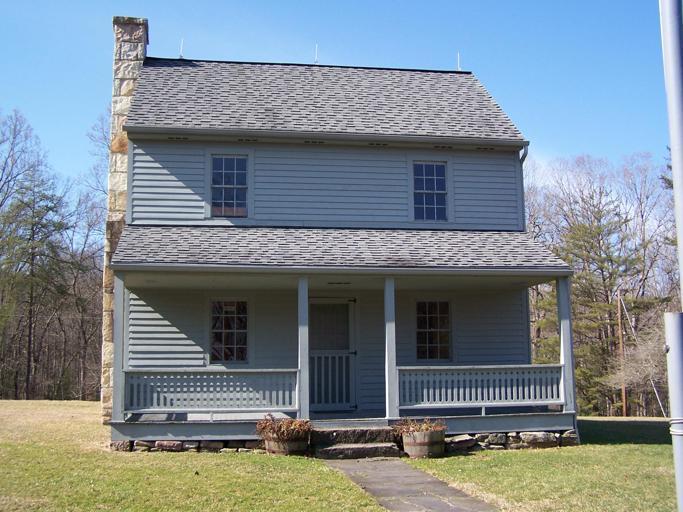}
}
\subfloat[Ground Truth]{
\includegraphics[width=0.25\linewidth]{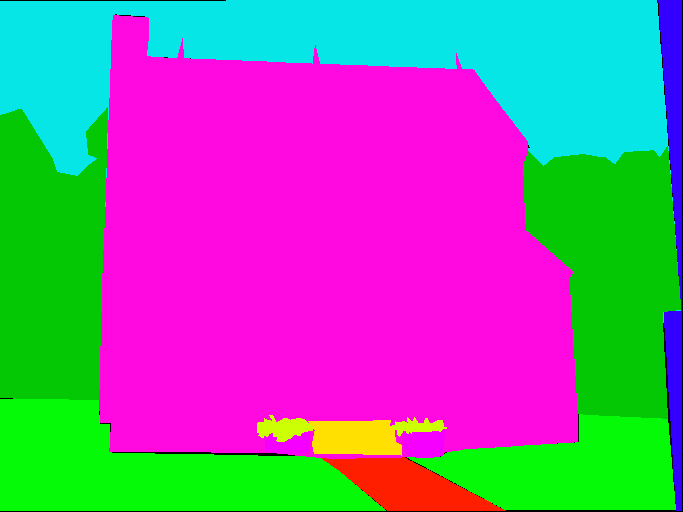}
}
\subfloat[FCN ({\small baseline})]{
\includegraphics[width=0.25\linewidth]{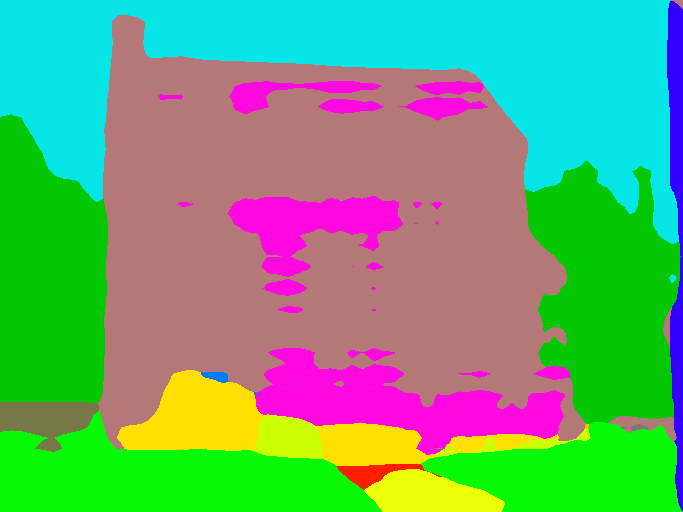}
}
\subfloat[EncNet ({\small \TextRed{ours}})]{
\includegraphics[width=0.25\linewidth]{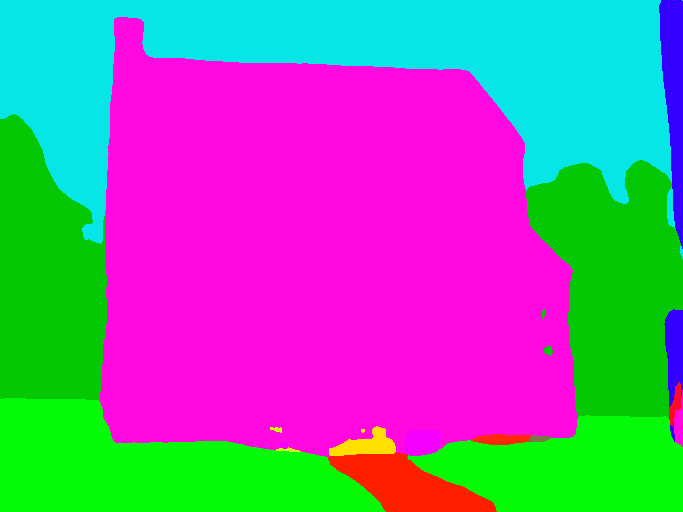}
}\\
\end{minipage}\hfill
\begin{minipage}[b]{0.13\linewidth}
\centering
\subfloat[Legend]{
\includegraphics[width=\linewidth,height=12cm]{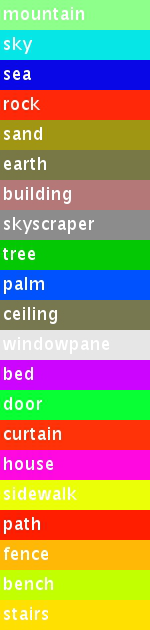}
}
\end{minipage}
\caption{Understanding contextual information of the scene is important for semantic segmentation. For example, baseline FCN classifies {\it sand} as {\it earth} without knowing the context as in 1$^{st}$ example. 
{\it building}, {\it house} and {\it skyscraper} are hard to distinguish without the semantics as in 2$^{nd}$ and 4$^{th}$ rows. In the 3$^{rd}$ example, FCN identify {\it windowpane} as {\it door} due to classifying isolated pixels without a global sense/view. (Visual examples from ADE20K dataset.)}
\label{fig:aderesults}
\vspace{-0.8em}
\end{figure*}

With the proposed Context Encoding Module, we build a Context Encoding Network (EncNet) with pre-trained ResNet~\cite{he2015deep}.
We follow the prior work using dilated network strategy on pre-trained network~\cite{chen2017rethinking,zhao2016pyramid,yu2017dilated} at stage 3 and 4\footnote{\label{ft:stage}We refer to the stage with original featuremap size 1/16  as stage 3 and size 1/32 as stage 4.}, as shown in Figure~\ref{fig:network}. 
We build our proposed Context Encoding Module on top of convolutional layers right before the final prediction, as shown in Figure~\ref{fig:overview}. 
For further improving the performance and regularizing the training of Context Encoding Module, we make a separate branch to minimize the SE-loss that takes the encoded semantics as input and predicts the presence of the object classes. 
As the Context Encoding Module and SE-loss are very light weight, we build another Context Encoding Module on top of stage 3 to minimize the SE-loss as an additional regularization, similar to but much cheaper than the auxiliary loss of PSPNet~\cite{zhao2016pyramid}. 
The ground truths of SE-loss are directly generated from the ground-truth segmentation mask without any additional annotations.  

Our Context Encoding Module is differentiable and inserted in the existing FCN pipeline without any extra training supervision or modification of the framework. 
In terms of computation, the proposed EncNet only introduces marginal extra computation to the original dilated FCN network.

\input{tables/ablation2}

\subsection{Relation to Other Approaches}

\paragraph{\bf Segmentation Approaches }

CNN has become  de facto  standard in computer vision tasks including semantic segmentation. The early approaches generate segmentation masks by classifying region proposals~\cite{girshick2014rich,hariharan2014simultaneous}. 
Fully Convolutional Neural Network (FCN) pioneered the era of end-to-end segmentation~\cite{long2015fully}. However, recovering detailed information from downsampled featuremaps is difficult due to the use of pre-trained networks that are originally designed for image classification. 
To address this difficulty, one way is to learn the upsampling filters, {\it i.e.}\ fractionally-strided convolution or decoders~\cite{badrinarayanan2015segnet,noh2015learning}. 
The other path is to employ Atrous/Dilated convolution strategy to the network~\cite{chen2014semantic,yu2015multi} which preserves the large receptive field and produces dense predictions. 
Prior work adopts dense CRF taking FCN outputs to refine the segmentation boundaries~\cite{chen2016Deeplab,cimpoi15}, and CRF-RNN achieves end-to-end learning of CRF with FCN~\cite{zheng2015conditional}. 
Recent FCN-based work dramatically boosts performance by increasing the receptive field with larger rate atrous convolution or global/pyramid pooling~\cite{chen2017rethinking,liu2015parsenet,zhao2016pyramid}. 
However, these strategies have to sacrifice the efficiency of the model, for example PSPNet~\cite{zhao2016pyramid} applies convolutions on flat featuremaps after Pyramid Pooling and upsampling and DeepLab~\cite{chen2016Deeplab} employs large rate atrous convolution that will degenerate to $1\times 1$ convolution in extreme cases. 
We propose the Context Encoding Module to efficiently leverage global context for semantic segmentation, which only requires marginal extra computation costs. 
In addition, the proposed Context Encoding Module as a simple CNN unit is compatible with all existing FCN-based approaches.

\paragraph{\bf Featuremap Attention and Scaling }
The strategy of channel-wise featuremap attention is inspired by some pioneering work. 
Spatial Transformer Network~\cite{jaderberg2015spatial} learns an in-network transformation conditional on the input which provides a spatial attention to the featuremaps without extra supervision.  
Batch Normalization~\cite{ioffe2015batch} makes the normalization of the data mean and variance over the mini-batch as part of the network, which successfully allows larger learning rate and makes the network less sensitive to the initialization method. 
Recent work in style transfer manipulates the featuremap mean and variance~\cite{dumoulin2016learned,huang2017adain} or second order statistics to enable in-network style switch~\cite{zhang2017multistyle}.
A very recent work SE-Net explores the cross channel information to learn a channel-wise attention and has achieved state-of-the-art performance in image classification~\cite{hu2017squeeze}. 
Inspired by these methods, we use encoded semantics to predict scaling factors of featuremap channels, which provides a mechanism to assign saliency by emphasizing or de-emphasizing individual  featuremaps conditioned on scene context.

\section{Experimental Results}
\begin{table}[t]
\centering
\resizebox{\linewidth}{!}
  {
  \begin{tabular} {l c c c c c | c c}
    \toprule[1pt]
    {\bf Method} & {\bf BaseNet} & {\bf Encoding} & {\bf SE-loss}  
    & {\bf MS} & {\bf pixAcc\%} & {\bf mIoU\%} \\
    \hline \hline
    FCN & Res50 & & & & 73.4 & 41.0 \\
    EncNet & Res50 & \checkmark & & & 78.1 & 47.6 \\
    EncNet & Res50 &\checkmark & \checkmark  & &  79.4 &  49.2 \\
    EncNet & Res101 &\checkmark  & \checkmark & &  80.4 &  51.7 \\
    EncNet & Res101 & \checkmark & \checkmark & \checkmark &  81.2 &  52.6 \\
    \bottomrule[1pt]
  \end{tabular}
  }
\caption{Ablation study on PASCAL-Context dataset. {\it Encoding} represents Context Encoding Module, {\it SE-loss} is the proposed Semantic Segmentation loss, 
{\it MS} means multi-size evaluation. Notably, applying Context Encoding Module only introduce marginal extra computation, but the performance is significantly improved. (PixAcc and mIoU calculated on 59 classes w/o background.) }
\vspace{-0.8em}
\label{tab:ablation}
\end{table}

In this section, we first provide implementation details for EncNet and baseline approach, 
then we conduct a complete ablation study on Pascal-Context dataset~\cite{mottaghi2014role}, and finally we report the performances 
on PASCAL VOC 2012~\cite{everingham2010pascal} and ADE20K~\cite{zhou2017scene} datasets. 
In addition to semantic segmentation, we also explore how the Context Encoding Module can improve the image classification performance of shallow network on CIFAR-10 dataset in Sec~\ref{exp:cifar}.

\subsection{Implementation Details}
\label{sec:detail}

Our experiment system including pre-trained models are based on open source toolbox PyTorch~\cite{pytorch}. We apply dilation strategy to {stage 3 and 4}\footnoteref{ft:stage} of the pre-trained networks with the output size of 1$/$8~\cite{chen2014semantic,yu2015multi}. 
The output predictions are upsampled  8 times using bilinear interpolation for calculating the loss~\cite{chen2017rethinking}. 
We follow prior work~\cite{zhao2016pyramid,chen2016Deeplab} to use the 
learning rate scheduling $lr=baselr * (1-\frac{iter}{total\_iter})^{power}$. The base learning rate is set to 0.01 for ADE20K dataset and 0.001 for others and the power is set to 0.9. The momentum is set to 0.9 and weight decay is set to 0.0001. 
The networks are training for 50 epochs on PASCAL-Context~\cite{mottaghi2014role} and PASCAL VOC 2012~\cite{everingham2010pascal}, and 120 epochs on ADE20K~\cite{zhou2017scene}. 
We randomly shuffle the training samples and discard the last mini-batch. 
For data augmentation, we randomly flip and scale the image between 0.5 to 2 and then randomly rotate the image between -10 to 10 degree and finally crop the image into fix size using zero padding if needed. 
For evaluation, we average the network prediction in multiple scales following~\cite{zhao2016pyramid,liu2015parsenet,schwartz2016material}.

In practice, larger crop size typically yields better performance for semantic segmentation, but also consumes larger GPU memory which leads to much smaller working batchsize for Batch Normalization~\cite{ioffe2015batch} and degrades the training. 
To address this difficulty, we implement Synchronized Cross-GPU Batch Normalization in PyTorch using NVIDIA CUDA \& NCCL toolkit, which increases the working batchsize to be global mini-batch size (discussed in Appendix~\ref{sec:app1}). 
We use the mini-batch size of 16 during the training. 
For comparison with our work, we use dilated ResNet FCN as baseline approaches. 
For training EncNet, we use the number of codewords 32 in Encoding Layers. 
The ground truth labels for SE-loss are generated by ``unique'' operation finding the categories presented in the given ground-truth segmentation mask.  
The final loss is given by a weighted sum of per-pixel segmentation loss and SE-Loss.

{\bf Evaluation Metrics } We use standard evaluation metrics of pixel accuracy (pixAcc) and mean Intersection of Union (mIoU). For object segmentation in PASCAL VOC 2012 dataset, we use the official evaluation server that calculates mIoU considering the background as one of the categories. For whole scene parsing datasets PASCAL-Context and ADE20K, we follow the standard competition benchmark~\cite{zhou2017scene} to calculate mIoU by ignoring background pixels. 


\subsection{Results on PASCAL-Context}
\begin{table}[t]
\centering
\resizebox{0.8\linewidth}{!}
  {
  \begin{tabular} {l c | c }
    \toprule[1pt]
    {\bf Method} & {\bf BaseNet} & {\bf mIoU\%} \\
    \hline \hline
    FCN-8s \cite{long2015fully} & & 37.8 \\
    CRF-RNN \cite{zheng2015conditional}&  & 39.3 \\
    ParseNet \cite{liu2015parsenet} & & 40.4 \\
    BoxSup \cite{dai2015boxsup} & & 40.5 \\
    HO\_CRF \cite{arnab2016higher} & & 41.3 \\
    Piecewise \cite{lin2016efficient} & & 43.3\\ 
    VeryDeep \cite{wu2016bridging} & & 44.5 \\
DeepLab-v2 \cite{chen2016Deeplab} & Res101-COCO & 45.7 \\ 
RefineNet \cite{lin2016refinenet} & Res152 &  \secbest 47.3\\
	\hdashline
    EncNet (\TextRed{\small ours}) & Res101 & \best 51.7 \\
    \bottomrule[1pt]
  \end{tabular}
  }
\caption{Segmentation results on PASCAL-Context dataset. (Note: mIoU on 60 classes w/ background.) }
\vspace{-0.8em}
\label{tab:pascalcontext}
\end{table}

PASCAL-Context dataset~\cite{mottaghi2014role} provides dense semantic labels for the whole scene, which has 4,998 images for training and 5105 for test. 
We follow the prior work~\cite{mottaghi2014role,chen2016Deeplab,lin2016refinenet} to use the semantic labels of the most frequent 59 object categories plus background (60 classes in total). We use the pixAcc and mIoU for 59 classes as evaluation metrics in the ablation study of EncNet. For comparing to prior work, we also report the mIoU using 60 classes in Table~\ref{tab:pascalcontext} (considering the background as one of the classes). 

\begin{figure}
\subfloat{
\includegraphics[width=0.24\linewidth]{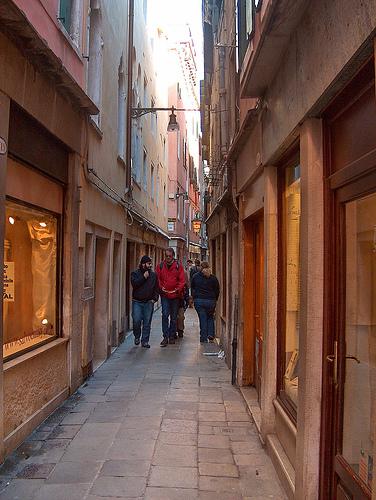}
}
\subfloat{
\includegraphics[width=0.24\linewidth]{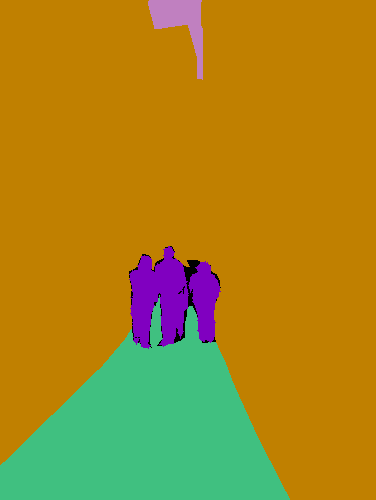}
}
\subfloat{
\includegraphics[width=0.24\linewidth]{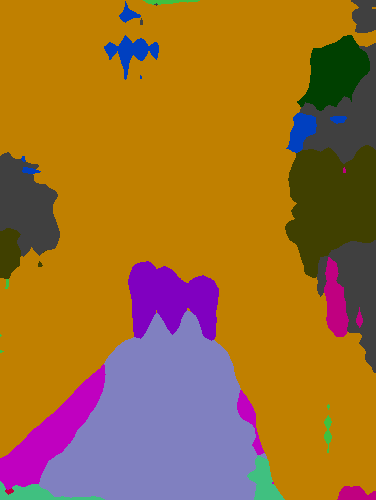}
}
\subfloat{
\includegraphics[width=0.24\linewidth]{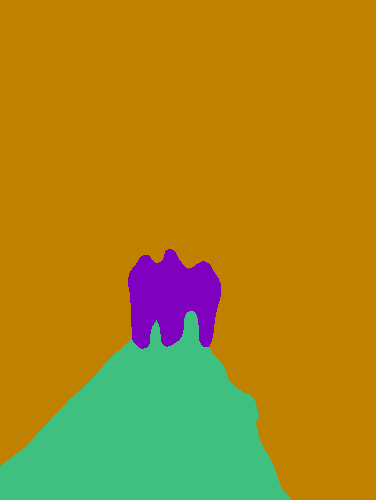}
}\par
\vspace{-0.8em}
\subfloat{
\includegraphics[width=0.24\linewidth]{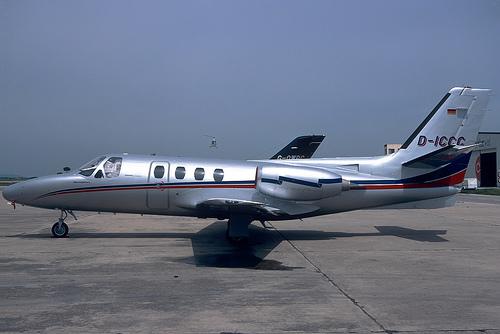}
}
\subfloat{
\includegraphics[width=0.24\linewidth]{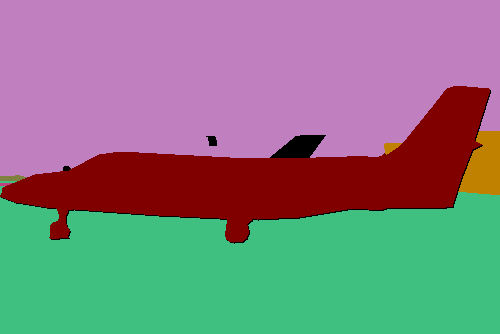}
}
\subfloat{
\includegraphics[width=0.24\linewidth]{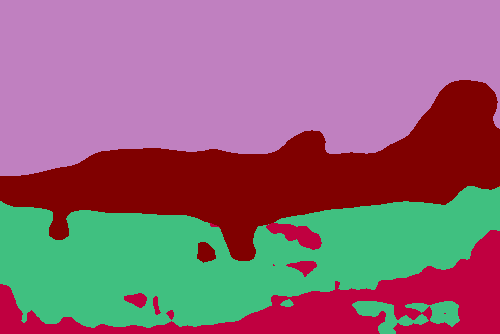}
}
\subfloat{
\includegraphics[width=0.24\linewidth]{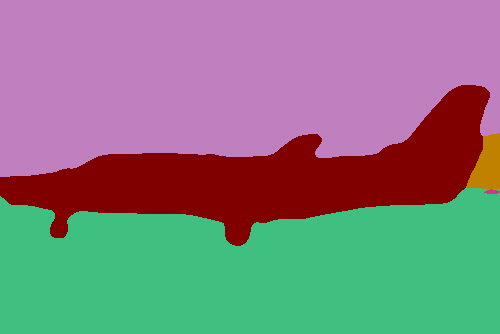}
}\par
\vspace{-0.8em}
\subfloat{
\includegraphics[width=0.24\linewidth]{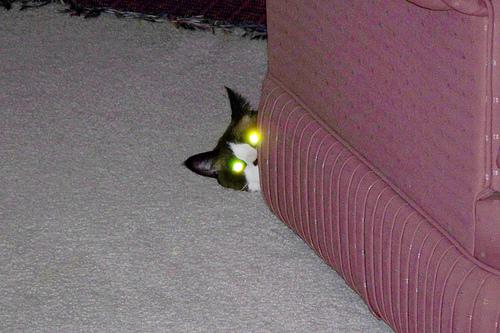}
}
\subfloat{
\includegraphics[width=0.24\linewidth]{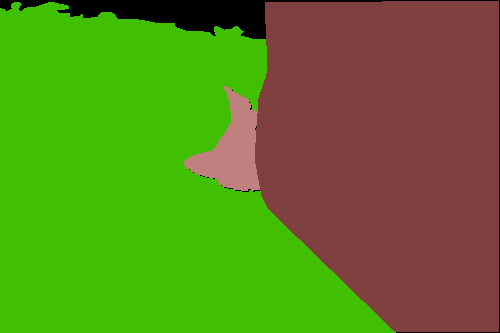}
}
\subfloat{
\includegraphics[width=0.24\linewidth]{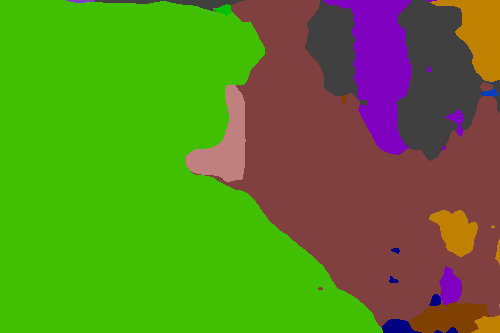}
}
\subfloat{
\includegraphics[width=0.24\linewidth]{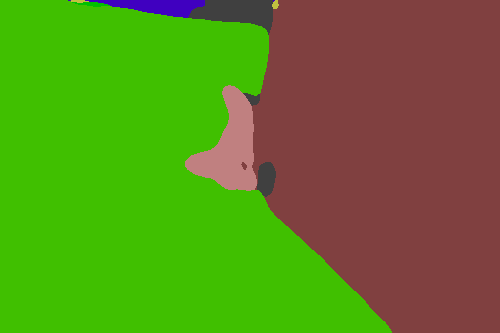}
}\par
\vspace{-0.8em}
\setcounter{subfigure}{0}
\subfloat[Image]{
\includegraphics[width=0.24\linewidth]{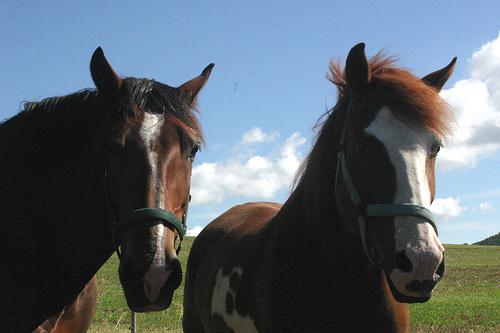}
}
\subfloat[Ground Truth]{
\includegraphics[width=0.24\linewidth]{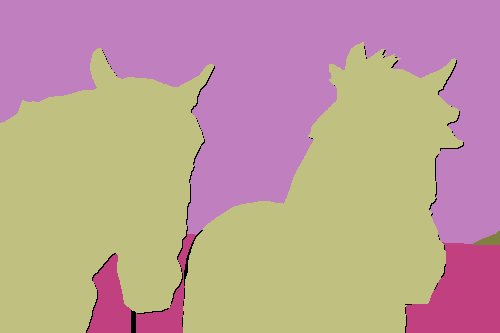}
}
\subfloat[FCN]{
\includegraphics[width=0.24\linewidth]{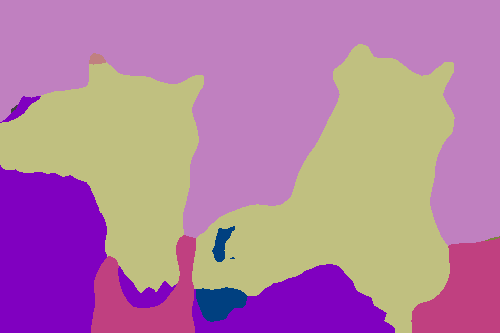}
}
\subfloat[EncNet (\TextRed{ours})]{
\includegraphics[width=0.24\linewidth]{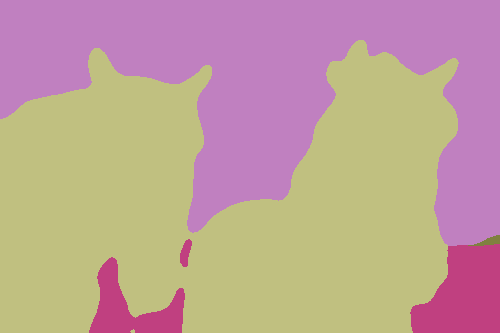}
}
\caption{Visual examples in PASCAL-Context dataset. EncNet produce more accurate predictions. }
\vspace{-0.8em}
\label{fig:pcontextv}
\end{figure}

\begin{table*}
    \footnotesize
    \setlength{\tabcolsep}{3pt}
    \begin{center}
    \begin{adjustbox}{max width=\textwidth}
        \begin{tabular}{ l | c c c c c c c c c c c c c c c c c c c c | c}
            \toprule[1pt]
            Method & aero & bike & bird & boat & bottle & bus & car & cat & chair & cow & table & dog & horse & mbike & person & plant & sheep & sofa & train & tv & mIoU \\
            \hline
            \hline
            FCN~\cite{long2015fully} & 76.8 & 34.2 & 68.9 & 49.4 & 60.3 & 75.3 & 74.7 & 77.6 & 21.4 & 62.5 & 46.8 & 71.8 & 63.9 & 76.5 & 73.9 & 45.2 & 72.4 & 37.4 & 70.9 & 55.1 & 62.2 \\
            DeepLabv2~\cite{chen2014semantic} & 84.4 & 54.5 & 81.5 & 63.6 & 65.9 & 85.1 & 79.1 & 83.4 & 30.7 & 74.1 & 59.8 & 79.0 & 76.1 & 83.2 & 80.8 & 59.7 & 82.2 & 50.4 & 73.1 & 63.7 & 71.6 \\
            CRF-RNN~\cite{zheng2015conditional} & 87.5 & 39.0 & 79.7 & 64.2 & 68.3 & 87.6 & 80.8 & 84.4 & 30.4 & 78.2 & 60.4 & 80.5 & 77.8 & 83.1 & 80.6 & 59.5 & 82.8 & 47.8 & 78.3 & 67.1 & 72.0 \\
            DeconvNet~\cite{noh2015learning} & 89.9 & 39.3 & 79.7 & 63.9 & 68.2 & 87.4 & 81.2 & 86.1 & 28.5 & 77.0 & 62.0 & 79.0 & 80.3 & 83.6 & 80.2 & 58.8 & 83.4 & 54.3 & 80.7 & 65.0 & 72.5 \\
            GCRF~\cite{vemulapalli2016gaussian} & 85.2 & 43.9 & 83.3 & 65.2 & 68.3 & 89.0 & 82.7 & 85.3 & 31.1 & 79.5 & 63.3 & 80.5 & 79.3 & 85.5 & 81.0 & 60.5 & 85.5 & 52.0 & 77.3 & 65.1 & 73.2 \\
            DPN~\cite{liu2015semantic} & 87.7 & 59.4 & 78.4 & 64.9 & 70.3 & 89.3 & 83.5 & 86.1 & 31.7 & 79.9 & 62.6 & 81.9 & 80.0 & 83.5 & 82.3 & 60.5 & 83.2 & 53.4 & 77.9 & 65.0 & 74.1 \\
            Piecewise~\cite{lin2016efficient} & 90.6 & 37.6 & 80.0 & 67.8 & 74.4 & 92.0 & 85.2 & 86.2 & 39.1 & 81.2 & 58.9 & 83.8 & 83.9 & 84.3 & 84.8 & 62.1 & 83.2 & 58.2 & 80.8 & 72.3 & 75.3 \\
            ResNet38\cite{wu2016wider} & \best 94.4 & \best 72.9 & \secbest{94.9} & 68.8 & \secbest{78.4} & 90.6 &\secbest{90.0} & 92.1 & \best 40.1 & 90.4 & \secbest 71.7 & 89.9 & \secbest 93.7 & \best{91.0} & \secbest 89.1 & \secbest 71.3 & \secbest 90.7 & \secbest 61.3 & \best 87.7 & \best{78.1} & 82.5\\
            PSPNet~\cite{zhao2016pyramid} & {91.8} & \secbest{71.9} & {94.7} & \secbest{71.2} & {75.8} & \secbest{95.2} & {89.9} & \best{95.9} & \secbest{39.3} & \best {90.7} & \secbest{71.7} & \best{90.5} & \best{94.5} & \secbest{88.8} & \best{89.6} & \best{72.8} & {89.6} & \best{64.0} & {85.1} & \secbest{76.3} & \secbest{82.6} \\
            \hdashline
            EncNet (\TextRed{\small ours})\footnoteref{fo:pascal} & \secbest 94.1 & 69.2 & \best 96.3 & \best 76.7 & \best 86.2 & \best 96.3 & \best 90.7 & \secbest 94.2 & 38.8 & \best 90.7 & \best 73.3 & \secbest 90.0 & 92.5 & \secbest 88.8 & 87.9 & 68.7 & \best 92.6 & 59.0 & \secbest 86.4 & 73.4 & \best 82.9 \\
            
            \hline
            \multicolumn{22}{c}{\textbf{With COCO Pre-training}}\\
            \hline
            
            CRF-RNN~\cite{zheng2015conditional} & 90.4 & 55.3 & 88.7 & 68.4 & 69.8 & 88.3 & 82.4 & 85.1 & 32.6 & 78.5 & 64.4 & 79.6 & 81.9 & 86.4 & 81.8 & 58.6 & 82.4 & 53.5 & 77.4 & 70.1 & 74.7 \\
            Dilation8~\cite{yu2015multi} & 91.7 & 39.6 & 87.8 & 63.1 & 71.8 & 89.7 & 82.9 & 89.8 & 37.2 & 84.0 & 63.0 & 83.3 & 89.0 & 83.8 & 85.1 & 56.8 & 87.6 & 56.0 & 80.2 & 64.7 & 75.3 \\
            DPN~\cite{liu2015semantic} & 89.0 & 61.6 & 87.7 & 66.8 & 74.7 & 91.2 & 84.3 & 87.6 & 36.5 & 86.3 & 66.1 & 84.4 & 87.8 & 85.6 & 85.4 & 63.6 & 87.3 & 61.3 & 79.4 & 66.4 & 77.5 \\
            Piecewise~\cite{lin2016efficient} & 94.1 & 40.7 & 84.1 & 67.8 & 75.9 & 93.4 & 84.3 & 88.4 & 42.5 & 86.4 & 64.7 & 85.4 & 89.0 & 85.8 & 86.0 & 67.5 & 90.2 & 63.8 & 80.9 & 73.0 & 78.0 \\
            DeepLabv2~\cite{chen2016Deeplab} & 92.6 & 60.4 & 91.6 & 63.4 & 76.3 & 95.0 & 88.4 & 92.6 & 32.7 & 88.5 & 67.6 & 89.6 & 92.1 & 87.0 & 87.4 & 63.3 & 88.3 & 60.0 & 86.8 & 74.5 & 79.7 \\
            RefineNet\cite{lin2016refinenet}  & 95.0 & 73.2 & 93.5 & 78.1 & 84.8 & 95.6 & 89.8 & 94.1 & 43.7 & 92.0 & 77.2 & 90.8 & 93.4 & 88.6 & 88.1 & 70.1 & 92.9 & 64.3 & 87.7 & 78.8 & 84.2\\
            ResNet38\cite{wu2016wider}  & \secbest 96.2 & 75.2 & \best 95.4 & 74.4 & 81.7 & 93.7 & 89.9 & 92.5 & \best 48.2 & 92.0 & 79.9 & 90.1 & 95.5 & \secbest 91.8 & 91.2 & \best 73.0 & 90.5 & 65.4 & 88.7 & 80.6 & 84.9\\
            PSPNet~\cite{zhao2016pyramid} & {95.8} & {72.7} & \secbest {95.0} & \secbest{78.9} & {84.4} & 94.7 & \best{92.0} & \secbest{95.7} & {43.1} & {91.0} & \best{80.3} & {91.3} & {96.3} & \best{92.3} & {90.1} & \secbest{71.5} & \best{94.4} & \secbest{66.9} & \secbest{88.8} & \best{82.0} & {85.4} \\
            DeepLabv3\cite{chen2017rethinking} & \best 96.4 & \secbest 76.6 & 92.7 & 77.8 & \best{87.6} & \best 96.7 & 90.2 & 95.4 &  47.5 & \secbest 93.4 & 76.3 & \secbest 91.4 & \best{97.2} &  91.0 & \best{92.1} & 71.3 & 90.9 & \best{68.9} & \best{90.8} & 79.3 & \secbest 85.7\\
            \hdashline
            EncNet (\TextRed{\small ours})\footnoteref{fo:pascal2} & 95.3 & \best 76.9 & 94.2 & \best 80.2 & \secbest 85.2 & \secbest 96.5 & \secbest 90.8 & \best 96.3 & \secbest 47.9 & \best 93.9 & \secbest 80.0 & \best 92.4 & \secbest 96.6 & 90.5 & \secbest 91.5 & 70.8 & \secbest 93.6 & 66.5 & 87.7 & \secbest 80.8 & \best 85.9\\
            \bottomrule[1pt]
        \end{tabular}
    \end{adjustbox}
    \end{center}
    \vspace{-1.5em}
    \caption{Per-class results on PASCAL VOC 2012 testing set. EncNet outperforms existing approaches and achieves 82.9\% and 85.9\% mIoU w/o and w/ pre-training on COCO dataset. (The best two entries in each columns are marked in gray color. Note: the entries using extra than COCO data are not included~\cite{luo2017deep,chen2017rethinking,wang2017learning}.)} 
\vspace{-0.8em}
\label{tab:pascal12}
\end{table*}

\paragraph{\bf Ablation Study. } To evaluate the performance of EncNet, we conduct experiments with different settings as shown in Table~\ref{tab:ablation}.  
Comparing to baseline FCN, simply adding a Context Encoding Module on top yields results of 78.1/47.6 (pixAcc and mIoU), which only introduces around 3\%-5\% extra computation but dramatically outperforms the baseline results of 73.4/41.0. 
To study the effect of SE-loss, we test different weights of SE-loss $\alpha=$\{0.0, 0.1, 0.2, 0.4, 0.8\}, and we find $\alpha=0.2$ yields the best performance as shown in Figure~\ref{fig:ablation2} (left). We also study effect of the number of codewords $K$ in Encoding Layer in Figure~\ref{fig:ablation2} (right), we use $K=32$ because the improvement gets saturated ($K=0$ means using global average pooling instead). 
Deeper pre-trained network provides better feature representations, EncNet gets additional 2.5\% improvement in mIoU employing ResNet101.
Finally, multi-size evaluation yields our final scores of 81.2\% pixAcc and 52.6\% mIoU, which is 51.7\% including background. Our proposed EncNet outperform previous state-of-the-art approaches~\cite{chen2016Deeplab,lin2016refinenet} without using COCO pre-training or deeper model (ResNet152) (see results in Table~\ref{tab:pascalcontext} and Figure~\ref{fig:pcontextv}).

\subsection{Results on PASCAL VOC 2012}

We also evaluate the performance of proposed EncNet on PASCAL VOC 2012 dataset~\cite{everingham2010pascal}, one of gold standard benchmarks for semantic segmentation. 
Following~\cite{long2015fully,dai2015boxsup,chen2017rethinking}, We use the augmented annotation set ~\cite{hariharan2015hypercolumns}, consisting of 10,582, 1,449 and 1,456 images in training, validation and test set. The models are trained on train+val set and then finetuned on the original PASCAL training set. 
EncNet has achieved 82.9\% mIoU\footnote{\label{fo:pascal}\tiny\bf\url{http://host.robots.ox.ac.uk:8080/anonymous/PCWIBH.html}} outperforming all previous work without COCO data and achieve superior performance in many categories, as shown in Table~\ref{tab:pascal12}. 
For comparison with state-of-the-art approaches, we follow the procedure of pre-training on MS-COCO dataset~\cite{lin2014microsoft}. 
From the training set of MS-COCO dataset, we select with images containing the 20 classes shared with PASCAL dataset with more than 1,000 labeled pixels, resulting in 6.5K images. All the other classes are marked as background. 
Our model is pre-trained using a base learning rate of 0.01 and then fine-tuned on PASCAL dataset using aforementioned setting. 
EncNet achieves the best result of 85.9\% mIoU\footnote{\label{fo:pascal2}\tiny\bf\url{http://host.robots.ox.ac.uk:8080/anonymous/RCC1CZ.html}} as shown in Table~\ref{tab:pascal12}. 
Comparing to state-of-the-art approaches of PSPNet~\cite{zhao2016pyramid} and DeepLabv3~\cite{chen2017rethinking}, the EncNet has less computation complexity. 

\begin{table}[t]
\centering
\resizebox{0.9\linewidth}{!}
  {
  \begin{tabular} {l c c | c }
    \toprule[1pt]
    {\bf Method} & {\bf BaseNet} & {\bf pixAcc\%} & {\bf mIoU\%} \\
    \hline \hline
    FCN \cite{long2015fully} & & 71.32 & 29.39 \\
SegNet \cite{badrinarayanan2015segnet} &  & 71.00 & 21.64 \\
DilatedNet \cite{yu2015multi} &  & 73.55 & 32.31 \\
CascadeNet \cite{zhou2017scene} &  & 74.52 & 34.90 \\
    RefineNet \cite{lin2016refinenet} & Res152 & - & 40.7 \\
    PSPNet \cite{zhao2016pyramid} & Res101 & 81.39 &  43.29 \\
    PSPNet \cite{zhao2016pyramid} & \secbest Res269 & \best 81.69 & \best 44.94 \\
	\hdashline
    FCN (\small baseline) & Res50 & 74.57 & 34.38 \\
    EncNet (\TextRed{\small ours}) & Res50 & 79.73 & 41.11 \\
    EncNet (\TextRed{\small ours}) & \best Res101 & \best 81.69 & \secbest 44.65 \\
    \bottomrule[1pt]
  \end{tabular}
  }
\caption{Segmentation results on ADE20K validation set. }
\vspace{-0.8em}
\label{tab:ade20k}
\end{table}

\begin{table}[t]
\centering
\resizebox{0.9\linewidth}{!}
{
  \begin{tabular} {c l | c }
    \toprule[1pt]
    {\bf rank} & {\bf Team} & {\bf Final Score} \\
    \hline
    - & (EncNet-101, single model \TextRed{\small ours}) & \best 0.5567\footnoteref{ft:adefinal} \\
    1 & CASIA\_IVA\_JD & 0.5547\\
    2 & WinterIsComing & 0.5544 \\
    - & (PSPNet-269, single model)~\cite{zhao2016pyramid} & \secbest 0.5538 \\
    \bottomrule[1pt]
  \end{tabular} 
}
  \caption{Result on ADE20K test set, ranks in COCO-Place challenge 2017. Our single model surpass PSP-Net-269 (1st place in 2016) and the winning entry of COCO-Place challenge 2017~\cite{zhou2017scene}.
}
\vspace{-0.8em}
\label{tab:coco}
\end{table}

\subsection{Results on ADE20K}

ADE20K dataset~\cite{zhou2017scene} is a recent scene parsing benchmark containing dense labels of 150 stuff/object category labels. The dataset includes 20K/2K/3K images for training, validation and set. We train our EncNet on the training set and evaluate it on the validation set using PixAcc and mIoU. 
Visual examples are shown in Figure~\ref{fig:aderesults}.  
The proposed EncNet significantly outperforms the baseline FCN. 
EncNet-101 achieves comparable results with  state-of-the-art PSPNet-269 using much shallower base network as shown in Table~\ref{tab:ade20k}. 
We fine-tune the EncNet-101 for additional 20 epochs on train-val set and submit the results on test set. 
The EncNet achieves a final score of 0.5567\footnote{\label{ft:adefinal}Evaluation provided by the ADE20K organizers.}, which surpass PSP-Net-269 (1st place in 2016) and all entries in COCO Place Challenge 2017 (shown in Table~\ref{tab:coco}). 


\subsection{Image Classification Results on CIFAR-10}
\label{exp:cifar}

In addition to semantic segmentation, we also conduct studies of Context Encoding Module for image recognition on 
CIFAR-10 dataset~\cite{cifar} consisting of 50K training images and 10K test images in 10 classes. 
State-of-the-art methods typically rely on very deep and large models~\cite{huang2016densely,he2015deep,he2016identity,xie2016aggregated}. In this section, we explore how much Context Encoding Module will improve the performance of a relatively shallow network, a 14-layer ResNet~\cite{he2015deep}. 
\vspace{-0.8em}

\paragraph{Implementation Details. } For comparison with our work, we first implement a wider version of pre-activation ResNet~\cite{he2016identity} and a recent work Squeeze-and-Excitation Networks (SE-Net)~\cite{hu2017squeeze} as our baseline approaches. 
ResNet consists a 3$\times$3 convolutional layer with 64 channels, followed by 3 stages with 2 basicblocks in each stage and ends up with a global average pooling and a 10-way fully-connected layer. The basicblock consists two 3$\times$3 convolutional layers with an identity shortcut. We downsample twice at stage 2 and 3, the featuremap channels are doubled when downsampling happens. 
We implement SE-Net~\cite{hu2017squeeze} by adding a Squeeze-and-Excitation unit on top of each basicblocks of ResNet (to form a SE-Block), which uses the cross channel information as a feedback loop. We follow the original paper using a reduction factor of 16 in SE-Block. 
For EncNet, we build Context Encoding Module on top of each basicblocks in ResNet, which uses the global context to predict the scaling factors of residuals to preserve the identity mapping along the network. For Context Encoding Module, we first use a 1$\times$1 convolutional layer to reduce the channels by 4 times, then apply Encoding Layer with concatenation of encoders and followed by a L2 normalization.

\begin{table}
   
\begin{center}
\begin{adjustbox}{width=0.8\linewidth}
\begin{tabular}{lccccc}
\toprule[1pt]
\multicolumn{1}{c}{Method} &\multicolumn{1}{c}{Depth} &\multicolumn{1}{c}{Params} &\multicolumn{1}{c}{Error} \\
\hline
\hline
ResNet (pre-act)~\cite{he2016identity}         &1001 &10.2M    &4.62 \\

Wide ResNet 28$\times$10 ~\cite{zagoruyko2016wide}     &28 &36.5M    &3.89  \\

ResNeXt-29 16$\times$64d~\cite{xie2016aggregated}  &29 &68.1M    &3.58   \\

DenseNet-BC (k=40)~\cite{huang2016densely}  &\secbest{190}    &25.6M    &\secbest 3.46 \\
\hdashline
ResNet 64d (baseline)    &14  &2.7M  &4.93 \\

Se-ResNet 64d (baseline)    &{14}  &2.8M &4.65 \\

EncNet 16k64d  (\TextRed{\small ours})        &\secbest{14}     &\best{3.5}M     &\secbest 3.96 \\

EncNet 32k128d (\TextRed{\small ours})        &\best{14}     &\secbest{16.8}M     &\best 3.45 \\
\bottomrule[1pt]
\end{tabular}
\end{adjustbox}
\end{center}
\vspace{-0.8em}
\caption{Comparison of model depth, number of parameters (M), test errors (\%) on CIFAR-10. $d$ denotes the dimensions/channels at network stage-1, and $k$ denotes number of codewords in Encoding Net.  }
\vspace{-0.8em}
\label{tab:cifar}
\end{table}

For training, we adopt the MSRA weight initialization~\cite{he2015delving} and use Batch Normalization~\cite{ioffe2015batch} with weighted layers. 
We use a weight decay of 0.0005 and momentum of 0.9. The models are trained with a mini-batch size of 128 on two GPUs  using a cosine learning rate scheduling~\cite{huang2016densely} for 600 epochs. 
We follow the standard data augmentation~\cite{he2015deep} for training, which pads the image by 4 pixels along each border and random crops into the size of 32$\times$32.
During the training of EncNet, we collect the statistics of the scaling factor of Encoding Layers $s_k$ and find it tends to be 0.5 with small variance. 
In practice, when applying a dropout~\cite{srivastava2014dropout}/shakeout~\cite{kang2016shakeout} like regularization to $s_k$ can improve the training to reach better optimum, by randomly assigning the scaling factors $s_k$ in Encoding Layer during the forward and backward passes of the training, drawing a uniform distribution between 0 and 1, and setting $s_k=0.5$ for evaluation. 


\input{tables/cifarbase}

We find our training process (larger training epochs with cosine lr schedule) is likely to improve the performance of all approaches. EncNet outperforms the baseline approaches with similar model complexity. 
The experimental results demonstrate that Context Encoding Module improves the feature representations of the network at an early stage using global context, which is hard to learn for a standard network architecture only consisting  convolutional layers, non-linearities and downsamplings.  
Our experiments shows that a shallow network of 14 layers with Context Encoding Module has achieved 3.45\% error rate on CIFAR10 dataset as shown in Table~\ref{tab:cifar}, which is comparable performance with state-of-the art approaches~\cite{huang2016densely,xie2016aggregated}.

\section{Conclusion}

To capture and utilize the contextual information for semantic segmentation, we introduce a Context Encoding Module, which selectively highlights the class-dependent featuremap and ``simplifies" the problem for the network. 
The proposed Context Encoding Module is conceptually straightforward, light-weight and compatible with existing FCN base approaches. 
The experimental results has demonstrated superior performance of the proposed EncNet. 
We hope the strategy of Context Encoding and our state-of-the-art implementation (including baselines, Synchronized Cross-GPU Batch Normalization and Encoding Layer) can be beneficial to  scene parsing and semantic segmentation work in the community.


\section*{Acknowledgement}
The authors would like to thank Sean Liu from Amazon Lab 126, Sheng Zha and Mu Li from Amazon AI for helpful discussions and comments.  
We thank Amazon Web Service (AWS) for providing free EC2 access. 


\appendix
\section*{Appendix}
\section{Implementation Details on Synchronized Cross-GPU Batch Normalization}
\label{sec:app1}
We implement synchronized cross-gpu batch normalization (SyncBN) on PyTorch~\cite{pytorch} using NVIDIA NCCL Toolkit. 
Concurrent work also implement SyncBN by first calculating the global mean and then the variance, which requires synchronizing twice in each iteration~\cite{liu2018path,peng2017megdet}. 
Instead, our implementation only requires synchronizing one time by applying a simple strategy:
for the $N$ number of given input samples $X=\{x_1,...x_N\}$, the variance can be represented by
\begin{equation}
\begin{split}
\sigma^2
&=\frac{\sum_{i=1}^N(x_i-\mu)^2}{N}\\
&=\frac{\sum_{i=1}^Nx_i^2}{N}-\frac{(\sum_{i=1}^Nx_i)^2}{N^2} ,
\end{split}
\label{eq:syncbn}
\end{equation}
where $\mu=\frac{\sum_{i=1}^Nx_i}{N}$. 
We first calculate $\sum x_i$ and $\sum x_i^2$ individually on each device, then the global sums are calculated by applying all reduce operation. The global mean and variance are calculated using Equation~\ref{eq:syncbn} and the normalization is performed for each sample $y_i=\gamma\frac{x_i-\mu}{\sqrt{\sigma^2+\epsilon}}+\beta$~\cite{ioffe2015batch}. Similarly, we synchronize once for the gradients of $\sum x_i$ and $\sum x_i^2$ during the back-propagation.

{\small
\bibliographystyle{ieee}
\bibliography{cvpr18}
}

\end{document}